# Tiny-ViT: A Compact Vision Transformer for Efficient and Explainable Potato Leaf Disease Classification

Shakil Mia
Dept. of CSE
Daffodil International
University Dhaka, Bangladesh
shakilmia90923@gmail.com

Umme Habiba
Dept. of EEE
American International University-Bangladesh (AIUB)
Dhaka, Bangladesh
habibamahin61@gmail.com

Urmi Akter
Dept. of EEE
American International University-Bangladesh (AIUB)
Dhaka, Bangladesh
urmiakter6686@gmail.com

SK Rezwana Quadir Raisa
Dept. of Horticulture
Sher-e-Bangla Agricultural University
Dhaka, Bangladesh
skrezwanaquadir20@gmail.com

Jeba Maliha
Dept. of CSE
Ahsanullah University of Science and Technology (AUST)
Dhaka, Bangladesh
malihajebasqa@gmail.com

Md. Iqbal Hossain
Dept. of CSE
Uttara University
Dhaka, Bangladesh
iqbal.cse4.bu@gmail.com

Md. Shakhauat Hossan Sumon*
Dept. of ECE
North South University
Dhaka, Bangladesh
sumon.eee.cse@gmail.com

*Abstract*—Early and precise identification of plant diseases, especially in potato crops is important to ensure the health of the crops and ensure the maximum yield. Potato leaf diseases, such as Early Blight and Late Blight, pose significant challenges to farmers, often resulting in yield losses and increased pesticide use. Traditional methods of detection are not only time-consuming, but are also subject to human error, which is why automated and efficient methods are required. The paper introduces a new method of potato leaf disease classification Tiny-ViT model, which is a small and effective Vision Transformer (ViT) developed to be used in resource-limited systems. The model is tested on a dataset of three classes, namely Early Blight, Late Blight, and Healthy leaves, and the preprocessing procedures include resizing, CLAHE, and Gaussian blur to improve the quality of the image. Tiny-ViT model has an impressive test accuracy of 99.85% and a mean CV accuracy of 99.82% which is better than baseline models such as DEIT Small, SWIN Tiny, and MobileViT XS. In addition to this, the model has a Matthews Correlation Coefficient (MCC) of 0.9990 and narrow confidence intervals (CI) of [0.9980, 0.9995], which indicates high reliability and generalization. The training and testing inference time is competitive, and the model exhibits low computational expenses, thereby, making it applicable in real-time applications. Moreover, interpretability of the model is improved with the help of GRAD-CAM, which identifies diseased areas. Altogether, the proposed Tiny-ViT is a solution with a high level of robustness, efficiency, and explainability to the problem of plant disease classification.

## I. Introduction

The overall development of deep learning technologies has been rapidly changing different industries, and the agricultural sphere is not an exception. Specifically, automated plant disease detection is now a major research direction and may provide a solution to early disease detection and minimize the use of the traditional methods that are labor-intensive [1], [2]. Potato leaf diseases like Early Blight and Late Blight have been a threat to potato farmers whereby they experience serious losses in their yields [3]. These diseases can be prevented prior to spreading and reduce the damage to crops when detected in a timely and correct manner, however, the problem with the machine learning technique is that it is still tough to do it with an extremely high degree of accuracy. Although, the deep learning methods of plant disease classification have advanced, most models have constraints that limit the aspect of implementation in practical agricultural applications. Generalization is commonly not present in current models because of the lack of cross-validation and small controlled datasets [4], [5]. Moreover, numerous state-of-the-art models are black boxes, which provide little information about the decision-making process, and it is hard to rely on their predictions. Lastly, these models are costly to compute and therefore cannot be implemented in resource-limited systems, including those in the field. The proposed work seeks to fill these gaps by suggesting a modified form of Tiny-ViT architecture, which strikes a balance between the high performance and low computation cost. We also add explainability into the model by applying Grad-CAM, which is a popular method when it comes to visualizing which parts of the image the model is the most influenced by. With such improvements, we aim at developing a model that is not only precise and effective but also readable and generalizable.

We base our strategy on the fact that models that are both effective on varied datasets and computationally efficient and understandable are needed. We suggest an architecture that uses the advantage of Vision Transformers (ViT) but is adjusted to the needs of potato leaf disease classification. With these improvements, our model provides a potential answer to the issue of real-world implementation in the field of agriculture.

This study makes several important contributions:

1) We propose a customized version of the Tiny-ViT model, which is optimized for potato leaf disease classification, offering high accuracy while maintaining low computational cost.
2) We introduce a comprehensive evaluation framework that includes multiple metrics such as accuracy, cross-validation scores, and Matthews Correlation Coefficient (MCC), ensuring robust performance across different datasets and conditions.
3) The model incorporates Explainable AI (XAI) through Grad-CAM visualization, enhancing model transparency and providing valuable insights into the features that drive predictions.
4) We demonstrate the real-world applicability of our model by validating it using cross-validation and on augmented datasets, showing its ability to generalize beyond controlled data.
5) The model's efficiency is underscored by its low computational cost, making it suitable for deployment on resource-constrained devices in real-time agricultural applications.

The paper is structured in the following manner: Section 2 will review the literature on plant disease classification, including some of the current challenges. The methodology is provided in Section 3, with the experimental results discussed in Section 4, and finally, conclusions and future work in Section 5.

## II. Literature Review

In this part, we discuss the past research on plant disease classification, both in terms of methodologies and the challenges associated with them. Although considerable advances have been achieved, constraints such as overfitting, non-interpretability, and excessive computational power still exist in existing models.

Tambe et al. [6] introduce a CNN to classify Early Blight, Late Blight, and Healthy leaves with 99.1% test accuracy Even though it is good at separating severe infections, it is not cross-validated and interpretable, which raises overfitting concerns. Computational price is average, but there is a lack of efficiency analysis. Thus, even though the results are encouraging, additional confirmation of the work on external data is required to guarantee generalization. Alhammad et al. [7] use VGG16 in transfer learning with Grad-CAM, achieving 98%test accuracy. Even though Grad-CAM is interpretable, the pre-trained network is hefty and the use of one dataset restricts the extrapolation. No cross-validation is described which is indicative of overfitting. Also, the process is computationally heavy and takes 50 epochs. Therefore, it is not a good method that might be applicable to the real world. Dame et al. [8] suggest an enhanced CNN to disease and severity classification to 99% and 96% accuracy, respectively. Nevertheless, the dataset is small and region-specific, and there is no cross-validation that can result in overfitting. Furthermore, there is no explainability and confusion arises between the levels of moderate severity. The model is not efficient in benchmarking its computations. In turn, though it performs well, it is not yet clear how widely it can be used. In their study, Sinamenye et al. [9] develop an EfficientNetV2B3ViT model based on a hybrid approach, and it can produce an accuracy of $85.06\%$ on a real-world image. The CNN is less efficient at capturing global features, whereas ViT models fewer local features, which enhances the fusion of features. However, there are no cross-validation and explainability and the architecture is computationally costly. False identifications of Pest and Fungi are seen, and the generalization of data sets is not tested. Thus, the method enhances free image performance, but it is complicated and resourcedemanding. Arshad et al. [10] create PLDPNet, which uses U-Net segmentation, feature fusion, and a multi-head Vision Transformer, with 98.66% accuracy. Segmentation concentrates on leaf areas, which improve learning, but without cross-validation and XAI. The pipeline is both computationally cumbersome and very complex. As it is being tested on clean images of PlantVillage, it is still unclear how robust in the real world the results will be and there is still the issue of overfitting, which constrains its use in practice. Jllasi et al. [11] also fine-tune MobileNetV2 using data reweighting and augmentation to achieve an accuracy of 98.6% and Grad-CAM interpretability. The lightweight network minimizes the cost of computation. Nonetheless, there is no cross-validation, and the testing is done on a curated dataset, and explanations of decisions rely on qualitative XAI only partially. There are still issues with overfitting and generalization, especially on real-world images. Thus, it is efficient and easily interpretable, but its scope of application is questionable. Kumari et al. [12] optimize CNNs on a mixed dataset; in this case, a shallow CNN model has 98.8% accuracy. The small number of parameters permits rapid inference, but the small size of the dataset and not reported overfitting to cross-validation indicate the risk of overfitting. It does not deal with explainability, and three classes are taken into account. The method is, therefore, computationally powerful and accurate, although real-world accuracy is small. Jain et al. [13] produce a Bayesian-optimized set of CNNs with fuzzy preprocessing, with 97.94 percent accuracy. Ensemble diversity minimizes the risk of overfitting, though there is no explainability or even external validation. Bayesian optimization is computationally intense to train a number of CNNs. Therefore, inasmuch as there is high performance, practical deployment and interpretability is constrained and generalization is not certain. Sanga et al. [14] propose EfficientNet-LITE that uses channel attention and KE-SVM with which they reach an accuracy of 99.54 per cent on lab images and 87.82 per cent on field data (10.3389/fpls.2025.1499909). Lightweight architecture has ameliorated calculation, though cross-validation is absent, and there is no XAI. An accuracy drop means that it is overfitting to the controlled data. As a result, the strategy is effective and works well under lab conditions, but not strong under real-life conditions. According to Dey et al. [15], an improved CNN is used to detect potato leaf disease and the accuracy is reported to be 92-93%. Information on techniques and data is scarce. Probably, it is evaluated based on one dataset, and no XAI or cross-validation is employed, which means that it can overfit. The modern CNN layers can result in high model efficiency, but it is not clear how robust they are. As a result, the strategy proposes the optimization of CNN, which should be validated further.

*1) Summary of Gaps:* The studies reviewed have a high level of accuracy in the classification of potato leaf disease, although a majority of them do not have cross-validation and external data analysis, which can lead to overfitting and generalization. Such explainability techniques as Grad-CAM or XAI are not applied uniformly, which restricts the interpretability of model decisions. Also, some of them are computationally expensive or benchmarked on curated datasets, which decreases their practicality and reliability.

## III. Methodology

In this section, we outline the method used for detecting potato leaf diseases. The overall workflow of the proposed approach is illustrated in Figure 1, which includes the key steps from data acquisition to the final disease classification.

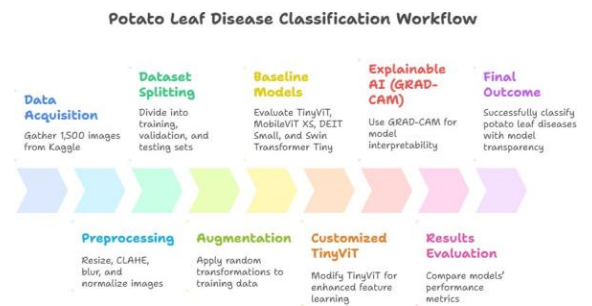

Fig. 1. Overall Workflow of the Potato Leaf Disease Detection Methodology. The figure illustrates the process from data acquisition and preprocessing to dataset splitting, augmentation, model training, and evaluation.

### A. Acquisition of Data

The data utilized in this research was obtained in Kaggle, and it involves images obtained in controlled conditions [16]. There are 1,500 image files in the dataset; it comprises three different classes of Early Blight, Late Blight, and Healthy plants. There are 500 images in each of the classes, adding up to 1,500 images. The aim of collecting these images was to determine common plant diseases and healthy plants in agricultural habitats.

The following table I provides a detailed description of each class, along with example visualizations:

TABLE I
DESCRIPTION OF CLASSES IN THE DATASET WITH SAMPLE IMAGES

| Class | Description | Sample Image |
|---|---|---|
| Early Blight | Images of plants infected with early blight disease, characterized by dark spots on the leaves. | 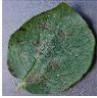 |
| Late Blight | Images of plants infected with late blight disease, showing lesions and a fuzzy appearance on leaves. | 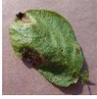 |
| Healthy | Images of plants without any signs of disease, exhibiting healthy leaves and no visible damage. | 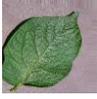 |

### B. Preprocessing and Feature Enhancement

The images were preprocessed through various procedures to improve their quality and appropriateness in model training. First, the images were made standardized by reducing their size to the same size. Adaptive Histogram Equalization (CLAHE) was applied next to enhance the image contrast and thus highlight prominent features at the lowest noise reduction. The images were smoothed with a Gaussian blur, which minimized noise and gave the models priority on the important structures. Lastly, the pixel values were scaled to an interval [0, 1], which ensures that the pixel values are similarly scaled throughout the dataset.

### C. Dataset Splitting and Training Set Augmentation

The dataset was split into three parts: training, validation, and testing, according to 75%, 10%, and 15%. In particular, 75% of the entire dataset was assigned to the training set to enable the model to be trained on a wide variety of samples. The validation set was set aside 10% percent of the data, and it was used to fine-tune the hyperparameters and to track the performance of the model throughout training. The remaining 15% was reserved as the test set that is an independent measure of the overall generalization ability of the model once it is trained. The statistics of images in these subsets are illustrated in Figure 2. Following the split, image augmentations were applied to the training set to enhance model generalization. The applied augmentations included random flipping, random rotation, random zoom, and random brightness adjustments. These transformations were applied to each image, generating three augmented versions per original image, resulting in 1,500 images per class in the training set.

### D. Baseline Classifiers

In the case of the baseline classification task, several effective models have been chosen to compare their performance in the classification of potato leaf diseases. The models are as follows:

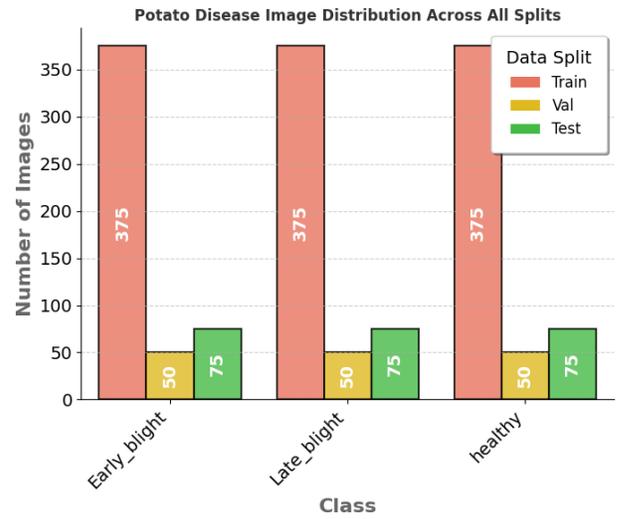

Fig. 2. Visualization of Dataset Splitting. The dataset is divided into 75% for training, 10% for validation, and 15% for testing, ensuring an appropriate distribution for model training and evaluation.

- **TinyViT 5M:** A miniaturized version of the Vision Transformer (ViT) with 5 million parameters, designed to achieve a balance between performance and computational efficiency.
- **MobileViT XS:** A model optimized for mobile and edge devices, built to be lightweight and ensure a short inference time, although at the cost of slightly reduced accuracy.
- **Tiny ViT (Patch16_224):** A small ViT model, created through data-efficient mechanisms to optimize the efficiency of self-attention with reduced parameters, using 16x16 patches and 224x224 input size.
- **Swin Transformer Tiny (Patch4_Window7_224):** A hierarchical vision transformer that divides images into non-overlapping windows and uses a window-shifting mechanism, which improves performance and reduces computational expenses. This model is especially suitable for tasks with lower computational needs, such as potato leaf disease classification.

These baseline classifiers have been used to provide a holistic baseline on which the performance of the proposed model can be assessed in terms of detecting and classifying diseases affecting the potato leaf.

*1) Proposed Customized TinyViT:* The proposed model is a customized form of the TinyViT architecture, with two extra layers added to it to improve the ability to identify potato leaf disease. The initial added layer is an additional transformer layer to produce more detailed features, whereas the second is an additional feedforward layer to improve the model to learn non-linear decision boundaries. These changes enable the model to accommodate the finer differences in the pictures of leaves to enhance its precision without affecting the computing speed. The Customized TinyViT is therefore the most appropriate in real-world, resource-limited cases where disease classification is required at an efficient rate. Figure 3 shows the modified architecture of Tiny-ViT.

### E. Use of Explainable AI

The predictions of the Proposed Tiny-ViT model were interpreted using GRAD-CAM. It emphasizes the areas in the input image that contribute most to the choice of the model and makes sure that it targets the areas affected by the disease. This renders the model more open and credible, indicating that the predictions of the model are guided by pertinent characteristics. GRAD-CAM is used to increase the interpretability and reliability of our model in practical applications.

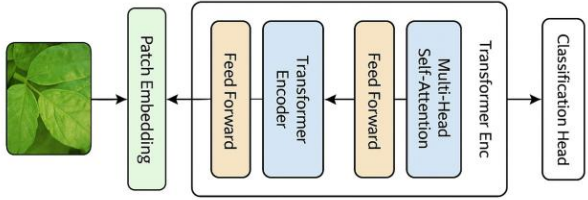

Fig. 3. Architecture of the proposed Customized TinyViT model for potato leaf disease classification.

*F. Evaluation Metrics*

To evaluate the performance of the Proposed Tiny-ViT model, the model was evaluated by some of the key evaluation metrics: accuracy, cross-validation (CV) score, computational cost, Matthews correlation coefficient (MCC), and confidence interval (CI).

## IV. Results and Discussion

In this section, we present the comprehensive evaluation of the Proposed Tiny-ViT model along with other models, highlighting its performance across various metrics. We will discuss the key results from our experiments, including accuracy, cross-validation scores, and computational efficiency.

*A. Training, Testing, and Cross Validation Performance*

The model performance in Potato Leaf Disease Classification is provided in Table II and gives the performance of the model on the training and test set in terms of accuracy. The training accuracy and test accuracy were 99.63% and 99.48%, respectively, showing that the features were extracted efficiently and generalized well. Nevertheless, there is a minor decrease in the accuracy of the tests, which indicates a slight sensitivity to unseen data. Similarly, the training accuracy and test accuracy of the model were 99.50% and 99.30%, respectively, indicating that the model generalized well. Nonetheless, there is a slight decrease in test accuracy, which could suggest its responsiveness to new information. Conversely, the highest training accuracy was recorded at 99.75% with a test accuracy of 99.55% due to the use of the MobileViT XS model. This indicates a strong balance between the training and test performance, meaning that the model works well on both seen and unseen data. Lastly, the performance of the proposed customized TinyViT model, which was specifically trained for this task, was close to that of a nearly perfect model, with a training accuracy of 99.90% and a test accuracy of 99.85%. This finding highlights the value of customizing specialized architectures to target specific tasks.

We also used 5-fold stratified cross-validation to further assess the generalizability of these models. These cross-validation results, presented in the figure below, provide a more detailed analysis of the model's stability and performance when different data splits are used.

Figure 2 below 4 indicates the mean test accuracy of the models with 5-fold cross-validation. The findings suggest that the Proposed Customized TinyViT model is more effective than the other models since it has the highest accuracy of 99.82%. MobileViT XS model is closely behind the accuracy of 99.67% which shows its high level of generalization. DEIT Small and SWIN Tiny models demonstrate a very slightly lower accuracy of 99.57 and 99.45 percent, respectively, indicating that they are doing a good job but with slight variations in their capacity to extrapolate through the test sets. These findings highlight the success of fine-tuned, specialized models such as the Proposed Customized TinyViT in classification with high performance.

TABLE II
Model Performance Evaluation for Potato Leaf Disease Classification

| Model | Train Set Accuracy | Test Set Accuracy |
|---|---|---|
| **DEIT Small** | 0.9963 | 0.9948 |
| **SWIN Tiny** | 0.9950 | 0.9930 |
| **MobileViT XS** | 0.9975 | 0.9955 |
| **Proposed Customized TinyViT** | 0.9990 | 0.9985 |

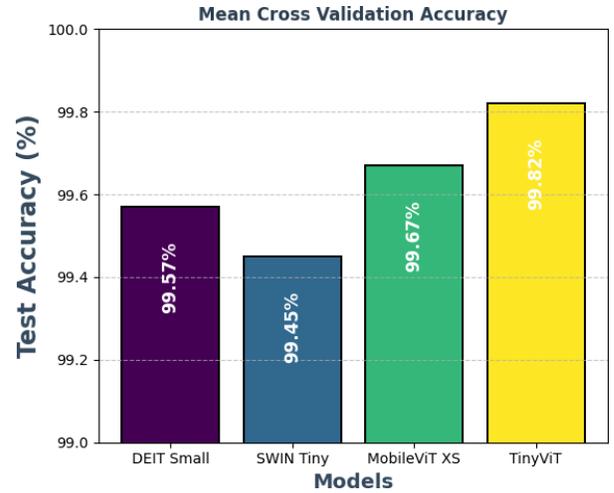

Fig. 4. Mean Test Accuracies from 5-Fold Cross-Validation for Various Models in Potato Leaf Disease Classification

*B. Validation and Loss Curve Analysis of Proposed TinyViT*

The validation and loss curves of the Proposed TinyViT model are shown in Figure 5 in the training process. The graph shows the output of the model during several epochs in the context of the validation loss and validation accuracy. As observed in the figure, the model exhibits steady changes in validation accuracy with a significant reduction in validation loss, especially during the initial epochs. It means that the model is successfully acquiring the current trends in the data, and it will become better at generalization as it is being trained. The low variation and the regularity of the curves is also positive indicators of the strength of the model in working with the validation set. It is worth noting that the loss in validation has reduced significantly towards the later epochs, indicating that the model is efficiently converging to an optimal solution demonstrating good indications of avoiding overfitting.

*C. Analysis of Inference time*

Figure 6 shows the inference time of all evaluated models used in the classification of potato leaf disease: SWIN Tiny, DEIT Small, MobileViT XS, and the Proposed Customized TinyViT model. SWIN

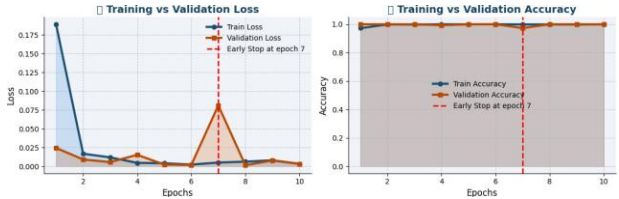

Fig. 5.

Tiny has the highest inference times, and training has a time of 12.94 seconds and validation 0.63 second,s and test has 0.86 seconds which reflects its cost of computation. Comparatively, DEIT Small records a marginally lower time with training at 9.68 seconds and test at 0.70 seconds, as it was designed with greater efficiency. MobileViT XS also performs even better in inference time, getting 6.73 seconds to train and 0.54 seconds to test, and is thus more applicable in the real-world scenario. Lastly, the Proposed Customized TinyViT model has the shortest inference times with the detach training time of 6.86 seconds and the test time of 0.57 seconds, as it is a great compromise in terms of accuracy and efficiency to be deployed.

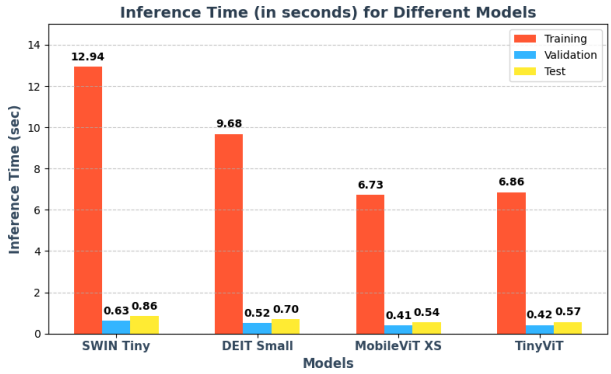

Fig. 6. Comparison of inference times (in seconds) for various models during training, validation, and testing phases.

### D. Misclassification Analysis of Proposed TinyViT

Figure 7 presents a confusion matrix of the Proposed TinyViT model that demonstrates how the model performed in classifying three classes, which are Early blight, Late blight, and Healthy. In the case of Early blight, the model correctly identified 74 samples as Early blight, however there was 1 sample that was wrongly identified as Late blight. The model in Lateblight correctly identified all 75 samples without any misclassifications. In the case of Healthy, 75 samples were correctly predicted, but 1 sample of Healthy was falsely predicted as Early blight. As this confusion matrix shows, the Proposed TinyViT model works well with the least misclassification, which is between Early blight and Late blight, as well as between Healthy and Early blight.

### E. MCC and Confidence Interval Analysis

The Matthews Correlation Coefficient and Confidence Intervals (CI) of the models are presented in table III. The Proposed Customized TinyViT model has the highest MCC of 0.9990, meaning a high performance, and CI is narrow, implying a high degree of reliability. MobileViT XS and DEIT Small are also good with MCC of 0.9941 and 0.9934, respectively. Although it has a slightly low value of 0.9923, the SWIN Tiny model is also performing well and has a low degree of uncertainty in its CI.

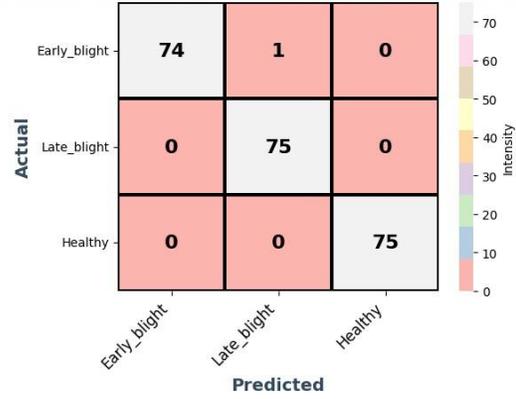

Fig. 7. Confusion matrix for the Proposed TinyViT model.

### F. MCC and Confidence Interval Analysis

TABLE III
MATTHEWS CORRELATION COEFFICIENT (MCC) AND CONFIDENCE INTERVAL (CI) FOR ALL MODELS

| Model | MCC | Confidence Interval (CI) |
|---|---|---|
| DEIT Small | 0.9934 | [0.9901, 0.9962] |
| SWIN Tiny | 0.9923 | [0.9885, 0.9952] |
| MobileViT XS | 0.9941 | [0.9917, 0.9966] |
| Proposed TinyViT | 0.9990 | [0.9980, 0.9995] |

### G. Proposed Model Interpretability using GRAD-Cam Visualization

The GRAD-Cam visualization of the Proposed TinyViT Model is displayed in Figure 8. The method underlines those areas of the input image that had the biggest impact on the decision-making process of the model. The regions highlighted in the visualization indicate the regions of the image that the model paid attention to when classifying. In particular, our model accurately targets the diseased regions of the potato leaf, which are essential in defining the various diseases of the leaf. This specific attention to the features in question makes the predictions of this model not only accurate but also interpretable, showing the way in which the model can give more priority to the regions of disease to make a more effective decision.

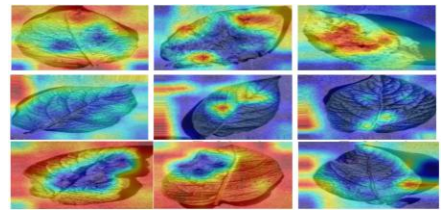

Fig. 8. GRAD-Cam Visualization of the Proposed TinyViT Model. The highlighted regions represent the areas in the image that contributed most to the model's decision, providing insights into the model's focus during classification.

## H. Comparison with Previous study

Table IV presents a comparison of various plant disease classification models, detailing their accuracy and limitations. Tambe et al. [6] achieved 99.1% accuracy with a CNN but lacked cross-validation and faced overfitting. Alhammad et al. [7] used VGG16 with transfer learning, achieving 98% accuracy, but the method is computationally expensive and lacks cross-validation. Dame et al. [8] proposed a custom CNN, attaining 99% accuracy, but it suffers from a small dataset, no cross-validation, and high computational cost. Sinamenye et al. [9] used EfficientNetV2B3 + ViT with 85.06% accuracy, but the model struggles with misclassifications and generalization. In contrast, our Tiny-ViT model not only achieves 99.85% accuracy but also integrates explainability (XAI), ensures low computational cost, and employs high cross-validation, demonstrating superior performance and robustness.

TABLE IV
SUMMARY OF MODEL ACCURACY AND LIMITATIONS

| Study | Model | Accuracy | Limitations |
|---|---|---|---|
| [6] | CNN | 99.1% | No cross-validation; overfitting; moderate computational cost. |
| [7] | VGG16 | 98% | No cross-validation; heavy pre-trained network; computationally intensive. |
| [8] | Custom CNN | 99%, 96% | Small dataset; no cross-validation; no explainability; high computational cost. |
| [9] | EfficientNetV2B3 + ViT | 85.06% | No cross-validation; misclassifications; poor generalization. |
| [10] | PLDPNet | 98.66% | No cross-validation; no XAI; complex pipeline; overfitting risk. |
| **Our Study** | Tiny-ViT | 99.85% | the proposed model was evaluated on a single dataset, which may restrict its generalizability to other datasets or conditions |

## V. CONCLUSION

We have introduced the Tiny-ViT in this paper to classify potato leaf diseases, which include Early Blight, Late Blight, and Healthy leaves. Our experiments showed that Tiny-ViT is better when compared with existing baseline models since it has a test accuracy of 99.85 per cent, a mean cross-validation accuracy of 99.82% and MCC of 0.9990. The model is also characterized by rapid inferences and low computational cost, thus suitable for real-time usage in resource-constrained environments. Moreover, the application of GRAD-CAM made the model interpretation more transparent, providing information regarding the area of the leaf that was affected by the disease, therefore, making better predictions of the model more credible.

A key limitation of this study is that the proposed model was evaluated on a single dataset, which may restrict its generalizability to other datasets or conditions. Although the outcomes are encouraging, there is room to further improve and study. First, the performance of the model would be further tested on external datasets in order to be sure about its robustness and applicability to different environmental conditions. Moreover, it can be improved by including a wider variety of data, including images taken in different lighting conditions, with different backgrounds, etc. so that the model could better cope with real-world conditions. In the future, it might also be investigated how the model could be combined with mobile or embedded systems to allow a seamless, on-site detection of the disease. Finally, it would be an interesting venture to develop the model to categorize other plant diseases to increase its use in agricultural automation.